\documentclass{article} % For LaTeX2e
\PassOptionsToPackage{numbers, compress}{natbib}
\usepackage[utf8]{inputenc}
\usepackage{graphicx}
\usepackage{tikz}
\usepackage{todonotes}
\usepackage{multirow}
\usepackage{amsmath}
\usepackage{cleveref}
\usepackage{subcaption}
\usepackage{multicol}
\usepackage{amssymb}
\usepackage{array}
\usepackage{bm}
\usepackage{enumitem}
\usepackage{algorithm}
\usepackage{algpseudocode}
\usepackage{tabularx}
\usepackage{bbm}
\usepackage{makecell}
\usepackage{colortbl}
\usepackage[normalem]{ulem}
\useunder{\uline}{\ul}{}
\usepackage{amsmath,amsfonts,bm}

\def\eqref#1{equation~\ref{#1}}
\def\1{\bm{1}}

\DeclareMathAlphabet{\mathsfit}{\encodingdefault}{\sfdefault}{m}{sl}
\SetMathAlphabet{\mathsfit}{bold}{\encodingdefault}{\sfdefault}{bx}{n}

\newcommand*\justify{%
  \fontdimen2\font=0.4em% interword space
  \fontdimen3\font=0.2em% interword stretch
  \fontdimen4\font=0.1em% interword shrink
  \fontdimen7\font=0.1em% extra space
  \hyphenchar\font=`\-% allowing hyphenation
}

\renewcommand{\texttt}[1]{%
  \begingroup
  \ttfamily
  \begingroup\lccode`~=`/\lowercase{\endgroup\def~}{/\discretionary{}{}{}}%
  \begingroup\lccode`~=`[\lowercase{\endgroup\def~}{[\discretionary{}{}{}}%
  \begingroup\lccode`~=`.\lowercase{\endgroup\def~}{.\discretionary{}{}{}}%
  \catcode`/=\active\catcode`[=\active\catcode`.=\active
  \justify\scantokens{#1\noexpand}%
  \endgroup
}

\usepackage{amsmath}
\usepackage{amssymb}
\usepackage{amsfonts}                               % blackboard math symbols
\usepackage{amsthm}
\usepackage[mathcal]{eucal}
\usepackage{mathrsfs}
\usepackage{bm}                                     % bm command
\usepackage{blkarray}                               % to support matrix
\usepackage{nicefrac}                               % compact symbols for 1/2, etc.

\usepackage{wrapfig}
\usepackage{graphicx}                               % include pdf figures
\usepackage{caption}
\usepackage{cleveref}
\usepackage{tikz}                                          
\usepackage{circuitikz}
\usetikzlibrary{patterns,snakes}
\usetikzlibrary{positioning,calc,fit,decorations.pathmorphing,shapes.geometric, shapes.gates.logic.US, calc}
\usetikzlibrary{arrows,arrows.meta,decorations.markings,shapes,shapes.arrows}
\usetikzlibrary{decorations,decorations.pathreplacing}
\usetikzlibrary{backgrounds}
\usepackage{filecontents}                           % support to pgfplots
\usepackage{pgfplots}
\usepackage{pgfplotstable}
\usepgfplotslibrary{groupplots}
\usepackage{scalefnt}
\pgfplotsset{compat=newest}
\usepackage{xcolor}
\usepackage{url}
\usepackage[preprint]{neurips_2024}

\definecolor{firstcolor}{HTML}{C3423F}
\definecolor{secondcolor}{HTML}{2A4B8C}

\DeclareMathOperator{\rank}{rank}
\DeclareMathOperator{\Norm}{norm}

\title{Every Token Counts: Generalizing 16M Ultra-Long Context in Large Language Models}

\author{
  \textbf{Xiang Hu$^{1*}$, 
  Zhanchao Zhou$^{1,2}$\thanks{~Equal contribution}, 
  Ruiqi Liang$^{1}$, 
  Zehuan Li$^{1}$, \\
  Wei Wu$^{1}$,
  Jianguo Li$^1$}\thanks{Corresponding author} \\
  $^1$ Ant Group \quad $^2$ Westlake University \\
  \texttt{\{aaron.hx, zhouzhanchao.zzc, liangruiqi.lrq, \\lizehuan.lzh, congyue.ww, lijg.zero\}@antgroup.com}
}

\begin{document}
\maketitle
%\clearpage
%\tableofcontents
%\newpage
%%%%%%%%%%%%%%%%%%%%%%%%%%%%%%%%%%%%%%%%%%%%%%%%%%%%%%%%%%%%%%%%%%
%%%%%%%%%%%%%%%%%%%%%%%%%%%%%%%%%%%%%%%%%%%%%%%%%%%%%%%%%%%%%%%%%%
%%%%%%%%%%%%%%%%%%%%%%%%%%%%%%%%%%%%%%%%%%%%%%%%%%%%%%%%%%%%%%%%%%
% main document
\begin{abstract}
This work explores the challenge of building “\textbf{Machines that Can Remember}”, framing long-term memory as the problem of efficient ultra-long context modeling. We argue that this requires three key properties: \textbf{sparsity}, \textbf{random-access flexibility}, and \textbf{length generalization}. To address ultra-long-context modeling, we leverage Hierarchical Sparse Attention (HSA), a novel attention mechanism that satisfies all three properties. We integrate HSA into Transformers to build HSA-UltraLong, which is an 8B-parameter MoE model trained on over 8 trillion tokens and is rigorously evaluated on different tasks with in-domain and out-of-domain context lengths to demonstrate its capability in handling ultra-long contexts.
Results show that our model performs comparably to full-attention baselines on in-domain lengths while achieving over 90\% accuracy on most in-context retrieval tasks with contexts up to 16M. This report outlines our experimental insights and open problems, contributing a foundation for future research in ultra-long context modeling. 
%Code is available at: \url{https://github.com/inclusionAI/Ling-Exp-UltraLong}.
\end{abstract}
\begin{figure}[h!]
    \centering
    \includegraphics[width=0.65\linewidth]{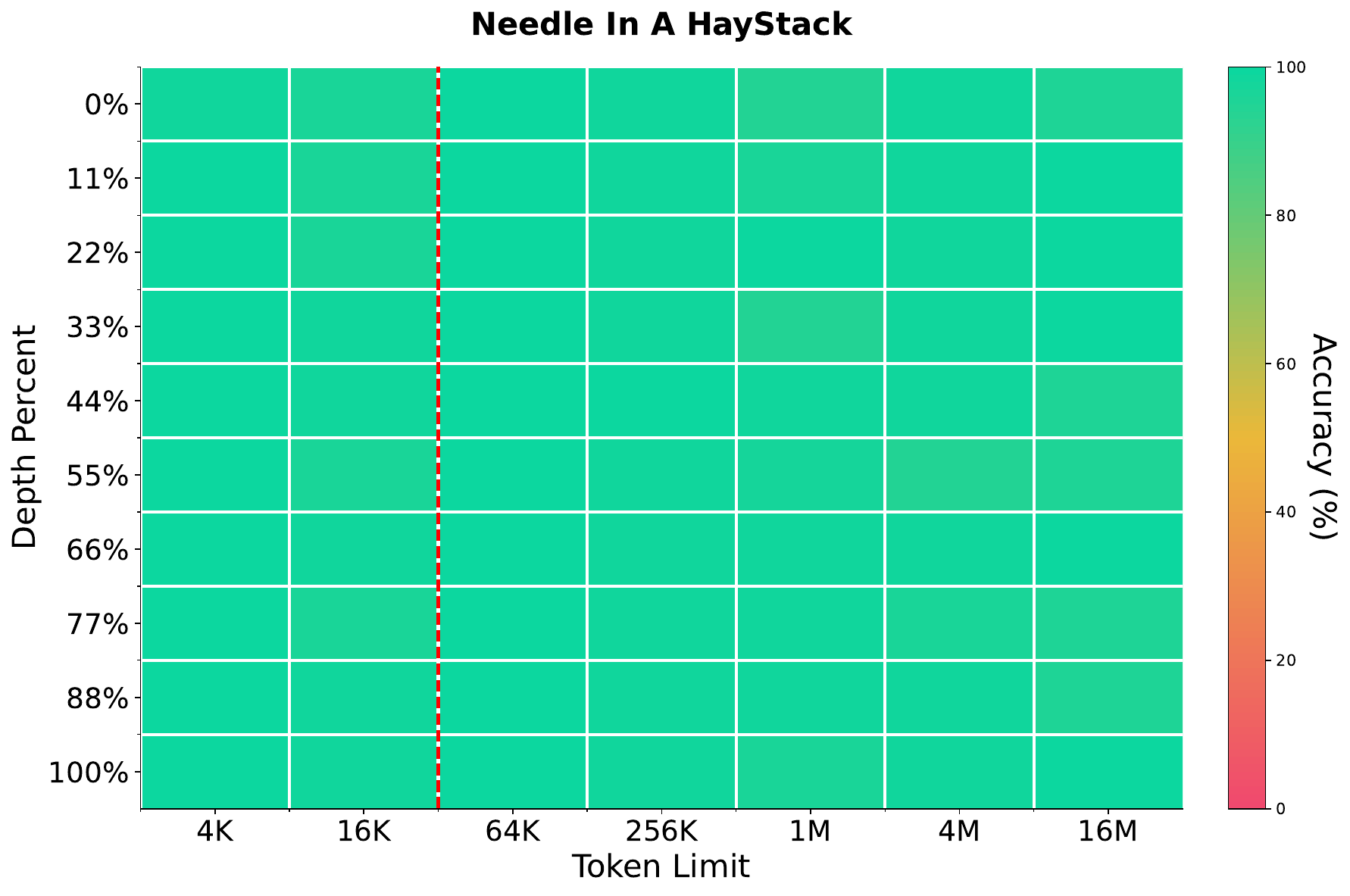} 
    \caption{Despite being pre-trained with an 8K context window and mid-trained up to 32K, HSA-UltraLong achieves near-perfect accuracy on S-NIAH even at a 16M-token context length. The red dashed line at 32K marks the boundary between in-domain (left) and out-of-domain (right).}
    \label{fig:main}
\end{figure}
\section{Introduction}
Despite the impressive capabilities of Large Language Models (LLMs)~\citep{DBLP:conf/nips/BrownMRSKDNSSAA20,achiam2023gpt,touvron2023llama}, their world knowledge is confined to static parameters, making it inflexible to update and impossible to learn dynamically from daily user interactions. This limitation motivates a fundamental question: how can we build machines that truly remember? Effective memory is critical for future AI agents, enabling each user to have a personalized agent that accumulates unique experiences over time. Human memory spans the entire context from birth to the present, suggesting that the problem of machine memory is closely related to ultralong context modeling.
Imagine if Transformers could efficiently handle infinite-length contexts—encompassing all pre-trained tokens—so that most world knowledge can be retrieved from context rather than compressed into model parameters. Furthermore, skills and the latest information could be acquired via in-context learning rather than through costly model retraining. Such advances would dramatically improve the online learning of knowledge and skills.

However, the Transformer~\citep{NIPS2017_3f5ee243} architecture, the backbone of modern LLMs, faces a fundamental efficiency challenge when processing ultra-long sequences, due to both poor length generalization and the quadratic computational complexity of full attention. Supporting longer contexts requires training models with extended context windows, yet simply scaling context length is computationally prohibitive.
To support ultra-long context, we argue that there are three key properties:
\begin{itemize}
    \item \textbf{Sparsity}:
    Human long-term memory operates through selective activation rather than full activation, retrieving relevant fragments as needed~\citep{Cowan2008WhatAT}. Clearly, using full attention to achieve an infinitely long context is not feasible. Therefore, implementing sparsity is one of the necessary conditions for ultra-long context modeling.
    \item \textbf{Random-Access Flexibility}:
    The effectiveness of sparsity relies on accurately retrieving relevant past information. Therefore, it is crucial to design an intrinsic retrieval mechanism within the model and optimize it end-to-end under the guidance of an auto-regressive loss.
    \item \textbf{Length Generalization}:
    Pretraining with an infinite context is impossible. To achieve the goal, the path must involve generalizing retrieval ability from short to long contexts.
\end{itemize}
While several approaches show promising paths to achieve the goal, each presents notable shortcomings. Recurrent architectures, such as Mamba~\citep{DBLP:journals/corr/abs-2312-00752,DBLP:conf/icml/DaoG24} and Linear Attentions~\citep{DBLP:conf/icml/KatharopoulosV020,DBLP:conf/iclr/YangKH25}, compress past variable-length information into a fixed-dimensional state vector. This introduces an information bottleneck and sacrifices random access to distant tokens. Similarly, sliding-window attention~\citep{Beltagy2020Longformer} suffers from the same fundamental constraint on distant context accessibility.
Sparse attention approaches like NSA~\citep{yuan-etal-2025-native} and MoBA~\citep{lu2025moba} improve training and inference efficiency over long sequences, but our empirical studies show they suffer from inaccurate chunk selection, which leads to both in-domain and out-of-domain performance degradation on in-context retrieval tasks. 

\begin{figure*}
    \centering
    \includegraphics[width=0.8\linewidth,trim={200pt 200pt 490pt 200pt},clip]{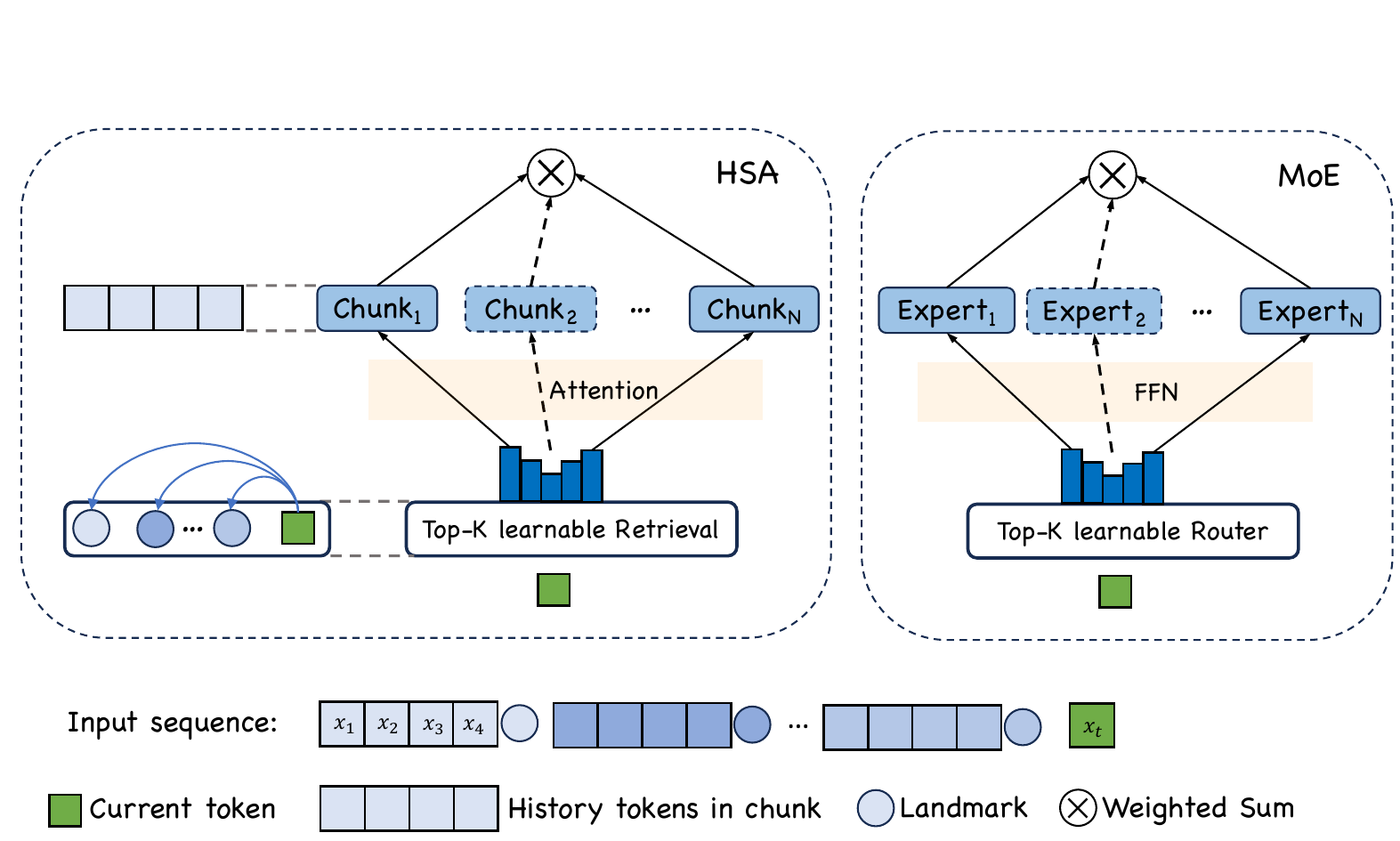}
    \caption{Hierarchical Sparse Attention (HSA) operates in a manner analogous to Mixture of Experts (MoE). First, the current token $x_t$ computes dot products with the landmark representations of past chunks as retrieval scores, from which the top-$k$ chunks are selected—similar to how MoE uses a router to select top-$k$ experts. Subsequently, $x_t$ performs attention with each of the $k$ retrieved chunks \textbf{separately}, mirroring the process in MoE where $x_t$ independently conducts Feedforward with $k$ experts. Finally, the attention outputs from each chunk are weighted by the softmax-normalized retrieval scores and summed, which is functionally equivalent to MoE’s fusion of outputs from the selected FFNs.}
    \vspace{-5mm}
    \label{fig:hierarchical_attn}
\end{figure*}

A recent line of work that combines model-inherent retrieval~\citep{mohtashami2023randomaccess,hu2025efficient} with chunk-wise sparse attention, such as Hierarchical Sparse Attention (HSA)~\citep{hu2025hardwarealigned}, has shown promising results in long-context modeling. Empirical studies~\citep{leng2025understandingimprovinglengthgeneralization} report that an HSA-based model pre-trained with a 4K context length can extrapolate to more than 10M context length while keeping high accuracy on the RULER~\citep{hsieh2024ruler} and BabiLong~\citep{kuratov2024babilongtestinglimitsllms} benchmark, which simultaneously satisfies \textbf{sparsity}, \textbf{random-access flexibility} and \textbf{length generalization}.
The method partitions text into fixed-length chunks with landmark representations; each token retrieves top-k relevant past chunks via these landmarks. The core innovation of HSA is to conduct attention with each chunk \textit{separately}, and then \textit{fuse the results weighted by the retrieval scores}. 
The overall process closely resembles the Mixture-of-Experts (MoE)~\citep{shazeer2017}, as illustrated in Figure~\ref{fig:hierarchical_attn}.
This design allows the retrieval scores to be integrated into the forward pass, enabling them to receive gradient updates during backpropagation. As a result, the model learns to assign higher retrieval scores to chunks that are more helpful for next token prediction. However, current work in this area is limited in scale and lacks results on data scaling, parameter scaling, and attention field scaling.

This work investigates the in-domain performance, extrapolation capability, and scaling challenges of HSA-based models, demonstrating a promising path toward effective ultra-long context modeling. We introduce HSA-UltraLong, an architecture that combines sliding-window attention with HSA, and train models from scratch across scales, including a 0.5B dense model and an 8B-A1B MoE model, on 8 trillion tokens. Training is followed by long-context extension and annealing. To preserve in-domain capability, we use a 4K-token sliding window. Our key findings include:
\begin{itemize}[itemsep=5pt, parsep=0pt]
    \item Effective length generalization requires the combination of chunk-wise attention, retrieval score–based fusion, and NoPE (No Positional Encoding); all three are essential.
    \item Sliding-window attention and HSA interact in nontrivial ways. HSA’s long-range generalization arises from learning to retrieve over short contexts and transferring that ability to long contexts. However, an overly large sliding window can weaken HSA’s learning of short-range dependencies, degrading generalization.
    \item The effective context length in the training corpus strongly influences length extrapolation.
\end{itemize}
We demonstrate successful extrapolation from a 32K pre-training window to 16M tokens trained on trillion-token-scale data. Our study presents a detailed analysis of in‑domain and extrapolation performance with respect to model size, attention window size, and data scaling, offering extensive experimental results and insights for future research on infinite‑length context modeling.

\section{Preliminary}
\subsection{Limitations of Chunk Selection in NSA}
\label{sec:nsa_ablation}
NSA is a highly inspiring contribution to sparse attention. However, our empirical study in Table~\ref{table:nsa_ablation} shows that its chunk selection mechanism does not always pick the most relevant chunk.
On the RULER benchmark~\citep{hsieh2024ruler}, NSA fails to achieve perfect accuracy even on in-domain tasks such as Multi-Query NIAH. We trace this to a key limitation: the chunk selection action is not end-to-end learnable. Further, our analysis of NSA’s extrapolation ability indicates that performance on in-context retrieval degrades rapidly as context length increases. Regarding positional encoding, we also find that No Positional Encoding (NoPE) supports extrapolation better than RoPE~\citep{DBLP:journals/ijon/SuALPBL24}.
\begin{table}[H]
\centering
\renewcommand{\arraystretch}{1.0}
\setlength\tabcolsep{6pt}
\small
\caption{NSA ablation with 4K as the in-domain length. The higher scores are shown in \textbf{bold}.}
\begin{tabularx}{\linewidth}{lc|ccccc|ccccc}
\toprule
\multirow{2}{*}{\textbf{Models}} & \multirow{2}{*}{\textbf{\#params}} & \multicolumn{5}{c|}{\textbf{Single-NIAH (ACC $\uparrow$)}} & \multicolumn{5}{c}{\textbf{MQ-NIAH(ACC $\uparrow$)}} \\
& & \makecell{\textbf{4K}} & \makecell{\textbf{8K}} & \makecell{\textbf{16K}} & \makecell{\textbf{32K}} & \makecell{\textbf{64K}}  & \makecell{\textbf{4K}} & \makecell{\textbf{8K}} & \makecell{\textbf{16K}} & \makecell{\textbf{32K}} & \makecell{\textbf{64K}} \\
\midrule
NSA(w/ RoPE)& 370M & 97.0&90.0&83.0&73.0&60.0&72.0&50.0&24.0&15.0&4.0\\
NSA(w/o RoPE)& 370M & \textbf{99.0}&\textbf{96.0}&\textbf{88.0}&\textbf{84.0}&\textbf{73.0}&\textbf{83.0}&\textbf{66.0}&\textbf{51.0}&\textbf{40.0}&\textbf{12.0}\\
\bottomrule
\end{tabularx}
\label{table:nsa_ablation}
\end{table}

\subsection{Attention with Chunk Retrieval}
As we mentioned, the challenge of sparse attention lies in accurately retrieving the previous chunks. 
Hierarchical Sparse Attention (HSA) addresses the challenge by jointly learning chunk selection and attention in an end-to-end manner.
Compared to NSA, HSA mainly makes two contributions:
\begin{itemize}
    \item \textbf{Retrieval-oriented sparse attention.} Specifically, each token conducts attention with each past chunk \textbf{separately} and then fuses the attention results via retrieval scores.
    \item \textbf{RoPE for short, NoPE for long.} To mitigate the negative impact of RoPE on extrapolation, the sliding-window attention's KV cache employs RoPE, while the HSA uses NoPE.
\end{itemize}

Formally, for an input sequence $\mathbf{S}=\{x_0, x_1,...,x_n\}$, where $n$ is the length of the sequence.
We denote the hidden states of tokens as $\mathbf{H} \in \mathbb{R}^{n \times d}$, where $d$ is the hidden dimension.
The whole sequence is split into chunks according to a fixed length $S$, which is set to 64 by default to better align with hardware, thus we have $\frac{n}{S}$ chunks in total. We use indices with $[\cdot]$ to indicate that it is indexed by chunk rather than by token, e.g., $\mathbf{H}_{[i]}:= \mathbf{H}_{iS:(i+1)S} \in \mathbb{R}^{S\times d}$. For each chunk, it has its own KV cache as $\mathbf{K}_{[i]}, \mathbf{V}_{[i]} \in \mathbb{R}^{S\times h \times d_{h}}$, with $h$ as the number of heads satisfying $h \times d_{h}=d$, and its landmark representation as $\mathbf{K}_i^{slc} \in \mathbb{R}^d$, which serves to summarize the content of the chunk. For each token, it uses $\mathbf{Q}^{slc}_t \in \mathbb{R}^d$ to retrieve chunks and $\mathbf{Q}^{attn}_t \in \mathbb{R}^{h \times d_h}$ to conduct attention with tokens inside chunks, both of which are derived from $\mathbf{H}_i$ via linear transformations. 
\begin{equation*}
s_{t,i}=\left\{\begin{matrix}
{\mathbf{Q}}_{t}^{slc\top} \mathbf{K}_{i}^{slc} / \sqrt{d},  & i \leq \lfloor\frac{t}{S}\rfloor\\
-\infty,    & i > \lfloor\frac{t}{S}\rfloor
\end{matrix}\right.\,,
\quad
\bm{\mathcal{I}}_t=\{i|\rank(s_{t,i})\ < K\},
\end{equation*}
where $\rank(\cdot)$ denotes the ranking position in descending order, and $\bm{\mathcal{I}}_t$ is the indices of $K$ chunks with the highest relevance scores for $x_t$.
\begin{gather}
\label{eq:hsa}
\mathbf{\bar{O}_{t,i}} = \text{Attention}(\mathbf{Q}_t^{attn}, {\mathbf{K}}_{[i]}, {\mathbf{V}}_{[i]}) = \underbrace{\text{Softmax}\left(\frac{\Norm(\mathbf{Q}_t^{attn}) \Norm({\mathbf{K}}_{[i]}^\top)}{\sqrt{d_h}}\right) {\mathbf{V}}_{[i]}}_{\textit{intra-chunk attention}}\,, \\
w_{t,i}=\frac{\exp(s_{t,i})}{\sum_{k \in \bm{\mathcal{I}}_t}\exp(s_{t,k})} \,,\quad
\mathbf{O_t} =\underbrace{\sum_{k \in \bm{\mathcal{I}}_t}{w_{t,k}\mathbf{\bar{O}}_{t,k}^l}}_{\textit{inter-chunk  fusion}}.
\end{gather}
$\Norm$ is the Query-Key Normalization~\citep{dehghani2023scalingvisiontransformers22,wortsman2023smallscale}, which we find to be very important for the stability of HSA in practical trillion-token scale training.

\section{Methodology}
In terms of model design, we use SWA for local information retrieval and HSA~\footnote{\url{https://github.com/ant-research/long-context-modeling}} for global information retrieval, fusing both local and global information through a stacking approach. A key challenge of long sequence inference is that the KV cache grows with the sequence length. Previous works~\citep{DBLP:conf/acl/WuT24,rubin-berant-2024-retrieval} have demonstrated that sharing the KV cache can significantly compress its size while maintaining comparable results. Inspired by these works, we share the intermediate layer KV cache among all HSA modules to serve as context memory.

\subsection{Model Architecture}
\begin{figure*}[tb!]
    \centering
    \includegraphics[width=0.6\linewidth]{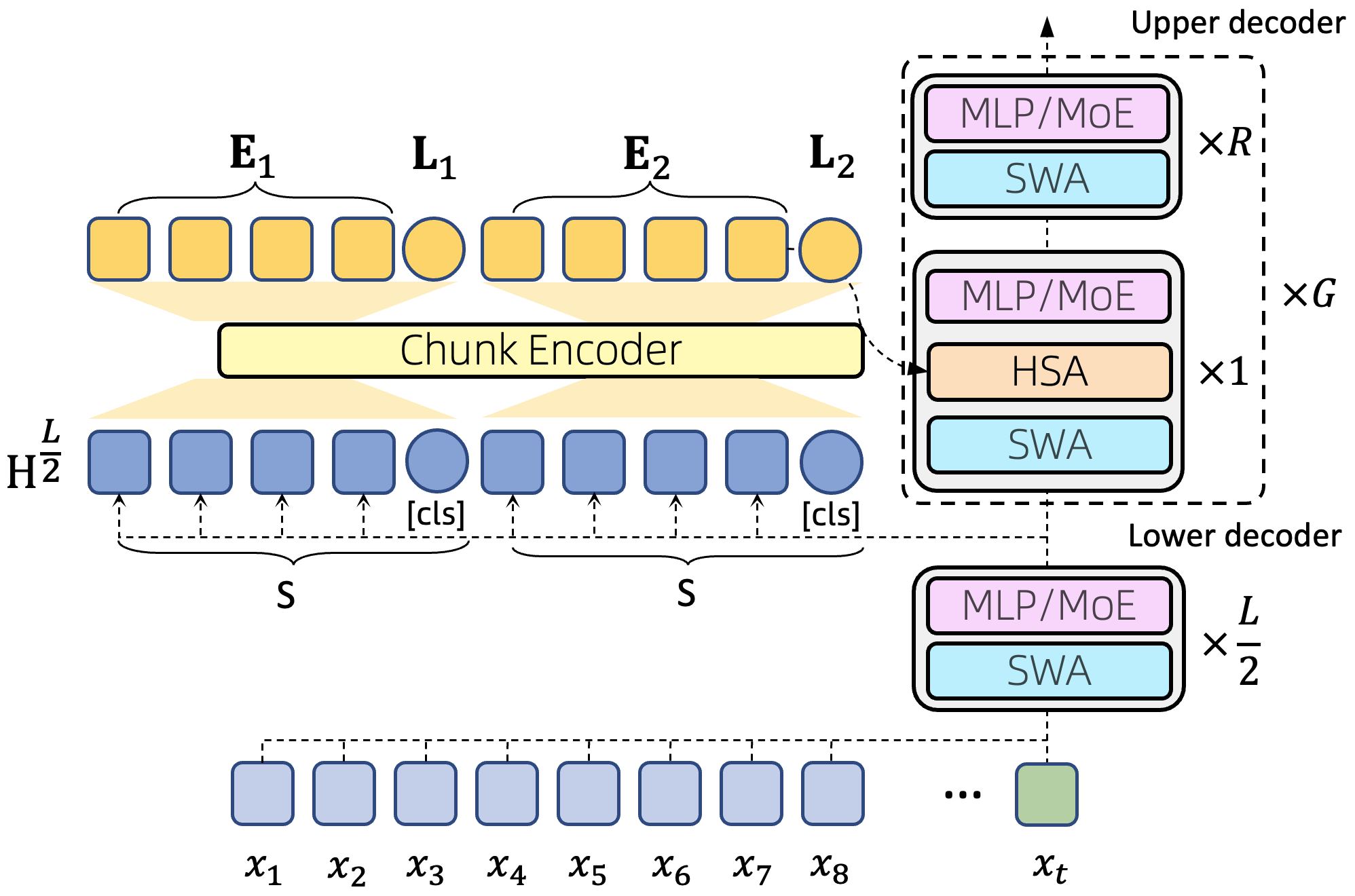}
    \caption{HSA-UltraLong model architecture.}
    \label{fig:model_arch}
\end{figure*}

Overall, as shown in Figure~\ref{fig:model_arch}, our model contains $L$ layers, which are divided into the upper decoder and the lower decoder. The lower decoder is composed of $\frac{L}{2}$ standard Transformer layers with SWA. For the upper decoder, we divide it into $G$ groups, where each group consists of one Transformer layer with both SWA and HSA, followed by several layers with SWA only.
For each chunk, we use a bi-directional encoder to obtain its summary representation. We denote $\mathbf{H}^l$ as the output hidden states of the $l$-{th} layer, and use the derivations from the intermediate layer output $\mathbf{H}^{\frac{L}{2}}$, as the chunk summary representation and KV cache, which are shared across all HSA modules. Each $\mathbf{H}^{\frac{L}{2}}_{[i]} \in \mathbb{R}^{S\times d}$ is accompanied by a [CLS] token, and is fed into a bi-directional encoder, as shown in Figure~\ref{fig:model_arch}, to obtain $\mathbf{E}_{[i]} \in \mathbb{R}^{S\times d}$ and $\mathbf{L}_i \in \mathbb{R}^d$. Finally, the keys and values used in Equation~\ref{eq:hsa} are obtained by applying a linear transformation to $\mathbf{E}_{[i]}$.
Regarding MoE, we follow the design of Ling-2.0~\citep{lingteam2025activationboostedscalinggeneral}, where the first layer of the model adopts a dense MLP structure and all subsequent layers use MoE. Each MoE block has one shared expert, following the design in DeepSeek V3~\citep{deepseekai2024deepseekv3}. We use training-free balance strategy~\citep{DBLP:journals/corr/abs-2408-15664} as the expert balancing strategy.

\subsection{Training}
Previous work~\citep{leng2025understandingimprovinglengthgeneralization} demonstrates that with a 512-token sliding window, HSA-based models pre-trained on a 4K context length can generalize to 32M on the RULER task with high accuracy. However, such a small sliding window often sacrifices downstream performance. To address this, we increase the sliding window size to 4K tokens. However, when training models from scratch with a 4K sliding window, we observe that \textbf{\textit{they fail to generalize beyond the 4K context length.}}
We hypothesize that HSA’s length generalization ability arises from its inherent retrieval mechanism, which learns to generalize from short to long contexts. An overly large sliding window can randomly access any short-range information, making it unnecessary for the HSA to focus on short-range patterns, thus preventing it from receiving meaningful gradients for short-context learning.
To overcome this limitation, we introduce a warmup stage before the pre-training phase. The whole pre-training procedure is as follows:
\begin{itemize}
    \item \textbf{Warm-up.} We use a short sliding window attention (SWA) of 512 tokens with a global HSA, setting top-k large enough to cover the full sequence. Synthetic ruler tasks are randomly inserted into 1\% of training samples. The warm-up phase is considered complete once the model achieves high needle-in-a-haystack retrieval accuracy on contexts well beyond the 512-token window. At this step, the context length is set to 16K.
    \item \textbf{Pre-training.} After the warm-up phase, we increase the SWA window size to 4K and decrease the HSA top-$k$, transitioning from dense to sparse attention. Continue training from the checkpoint at the end of warm-up. The context length remains 16K.
    \item \textbf{Long-context mid-training.}
    Switch to corpora with longer effective contexts and raise HSA top-k to cover the full sequence. The context length is expanded to 32K.
    \item \textbf{Annealing.} Perform annealing on high-quality data while keeping a 32K context length.
    \item \textbf{Supervised fine-tuning.} Perform supervised fine-tuning (SFT) with an 8K context length.
\end{itemize}
\section{Experiments}
\paragraph{Training Data}
In the first phase of general pretraining, we utilized a large-scale deduplicated, multi-domain dataset totaling 10T tokens, with differential sampling ratios across various sub-datasets from different domains. This distribution comprised predominantly Web content (50\%), followed by Code (14.4\%), Math (12.0\%), Code-nlp (5.6\%), Reason (5\%), Multilingual (4.0\%), Books (2.0\%), Wikipedia (1.5\%), and Others (5.5\%). During this phase, the MoE model processed 8T tokens while the dense model was trained on 4T tokens. The second phase uses a dataset of 32K-length long-text sequences totaling 175B tokens and the third phase consisted of 400B tokens with a high proportion of reasoning data. During the Supervised Fine-tuning phase, we utilized the same dataset as described in \cite{wu2025grove}.

\paragraph{Hyperparameters}
All models are trained using AdamW optimizer with a weight decay of 0.01, $\beta_1 = 0.9$, $\beta_2 = 0.95$, and gradient clipping norm of 1.0. We use FSDP2 for distributed training. For the MoE model, we employ a learning rate of 3.87e-4, sequence length of 16,384, and batch size of 16.8M tokens. The dense model is trained with a learning rate of 4.96e-4 and batch size of 5.2M tokens. The learning rate schedule begins with a linear warmup phase followed by a constant learning rate maintained until training completion. For the Supervised Fine-tuning stage, we adopt a cosine decay learning rate schedule. The dense model uses a learning rate of 5.5e-5 and was trained for up to 5 epochs, while the MoE model uses a learning rate of 3.87e-4 and was trained for up to 3 epochs. In both cases, we select the checkpoint from the epoch that yields the best performance.

\paragraph{Evaluation Benchmarks}
To conduct a comprehensive evaluation of the model, we selected a diverse range of assessment tasks, encompassing four major categories: general tasks, mathematical tasks, coding tasks, and alignment tasks:
\begin{itemize}
    \item \textbf{General Tasks:} MMLU~\citep{hendrycks2020mmlu}, CMMLU~\citep{li2023cmmlu}, C-Eval~\citep{huang2023ceval}, ARC~\citep{arc}, AGIEval~\citep{agieval}, PIQA~\citep{piqa}, HellaSwag~\citep{hellaswag} and BBH~\citep{suzgun2022bbh}.
    \item \textbf{Math Tasks:} GSM8K~\citep{cobbe2021gsm8k}, MATH~\citep{hendrycks2021math}, CMATH~\citep{cmath}, MATH-500~\citep{lightman2023math500} and OlympiadBench~\citep{he2024olympiadbench}.
    \item \textbf{Coding Tasks:} HumanEval~\citep{humaneval}, HumanEval+~\citep{liu2023evalplus}, MBPP~\citep{mbpp}, MBPP+~\citep{liu2023evalplus}, and CRUX-O~\citep{gu2024cruxeval}.
    \item \textbf{Alignment Tasks:} We report the average prompt-level strict accuracy of IFEval~\citep{zhou2023ifeval}
\end{itemize}

\subsection{Small-scale Preliminary Experiments}
\begin{table}[t]
\centering
\renewcommand{\arraystretch}{1.0}
\setlength\tabcolsep{6.5pt}
\small
\caption{Preliminary experiments on HSA-UltraLong-Base with a training context length of 16k tokens. The highest and second-best scores are shown in \textbf{bold} and \underline{underlined}, respectively.}
\begin{tabularx}{\linewidth}{lcc|ccc|cccc}
\toprule
\multirow{2}{*}{\textbf{Models}} & \multirow{2}{*}{\textbf{\#params}} & \multirow{2}{*}{\textbf{Warmup}} & \multicolumn{3}{c|}{\textbf{PG19 (PPL $\downarrow$)}} & \multicolumn{4}{c}{\textbf{MQ-NIAH(ACC $\uparrow$)}} \\
& & & \makecell{\textbf{4K}} & \makecell{\textbf{8K}} & \makecell{\textbf{16K}}  & 
\makecell{\textbf{4K}} & \makecell{\textbf{8K}} & 
\makecell{\textbf{64K}} & \makecell{\textbf{1M}} \\
\midrule
BaseLM & 519.6M & - & \underline{18.61} & 17.53 & 16.77 & 89.0& 23.0& 5.0 & 0.0 \\
SWA+HSA & 537.7M & self-copy & 18.87 & \underline{17.44} & \underline{16.50} & \textbf{100.0} & \textbf{96.0} & \textbf{93.0} & \textbf{93.0} \\
SWA+HSA & 537.7M & short-swa,full-hsa &  \textbf{18.30} & \textbf{17.13} & \textbf{15.96} & \underline{99.0} & \underline{95.0} & \underline{90.0} & \underline{66.0}  \\
\bottomrule
\end{tabularx}
\label{tab:prelimilary_hsa}
\end{table}
In this section, we detail the architecture of HSA-UltraLong and validate its ability to balance in-domain performance with length extrapolation capabilities through small-scale experiments. We compared our model against Base Language Model (BaseLM), which uses full attention across all layers. HSA-UltraLong employs SWA with a window size of 4K across all layers, while strategically replacing two layers with HSA layers. The HSA layers maintain a SWA window size of 512 and, following our findings in Section~\ref{sec:nsa_ablation}, we removed positional encoding information from these layers. Each HSA layer includes an additional encoder sub-layer, resulting in less than 5\% parameter increase compared to BaseLM. Within the HSA layers, we set the chunk size to 64 and top-k to 64, establishing a fixed historical context window of 4,096 tokens.

Our initial experiments revealed limited length extrapolation capabilities when training the model directly with such configuration. We hypothesized that this limitation stemmed from the predominance of pretraining data requiring only short-range modeling capabilities within the 4K window of SWA layers, leaving the HSA modules insufficiently trained. To address this issue, we explored two warm-up strategies:

\begin{itemize}[itemsep=5pt, parsep=0pt]
    \item \textbf{Self-copy warm-up:} 
    We keep the model architecture unchanged and initialize training with a self-copy objective. Given an input sequence $\mathbf{S} = \{x_1, …, x_n\}$, we construct a target sequence $\mathbf{S}' = \{x_1, …, x_n, x_1, …, x_n\}$ by concatenating $\mathbf{S}$ with itself. This objective encourages the model to attend to and retrieve long-range prefix information, enabling it to reconstruct the second half of the sequence.
    \item \textbf{Full HSA + Short SWA warm-up:} Setting top-k in HSA layers to 256 and sliding window size to 512 during the initial training phase.
\end{itemize}

All experiments were conducted on a 0.5B parameter dense model trained on 100B tokens with a pre-training context length of 16k. We incorporated 1\% ruler-specific synthetic data into the pre-training data to facilitate evaluation using ruler benchmarks. Performance was evaluated based on the perplexity of the last 4k tokens on the PG19 dataset and accuracy on the Multi-key NIAH (MK-NIAH) task within the ruler benchmark.

The results in Table~\ref{tab:prelimilary_hsa} demonstrated that the self-copy warm-up strategy yielded the best length extrapolation performance, albeit with some negative impact on in-domain performance. The full HSA + short SWA warm-up approach achieved a better balance, maintaining in-domain performance while delivering reasonable length extrapolation capabilities.

\subsection{Scaling and Evaluation of Pretrained and Fine-tuned Models}
\begin{table}[t]
\centering
\renewcommand{\arraystretch}{1.0}
\setlength\tabcolsep{6.5pt}
\small
\caption{Comparison among HSA-UltraLong-Base (HSA-UL-Base) and other baselines. All models were evaluated under a unified framework for fair comparison.}
\begin{tabularx}{\linewidth}{lccc|ccc}
\toprule
& \makecell{\textbf{Qwen2.5} \\ \textbf{Annealing}} & \makecell{\textbf{Qwen3} \\ \textbf{Annealing}} & \makecell{\textbf{HSA-UL} \\ \textbf{Annealing}} & \makecell{\textbf{TRM-MoE} \\ \textbf{Base}} & 
\makecell{\textbf{HSA-UL} \\ \textbf{Base}} & 
\makecell{\textbf{HSA-UL} \\ \textbf{Annealing}} \\
\midrule
Architecture & Dense & Dense & Dense & MoE & MoE & MoE \\
\# Total Params & 0.5B & 0.6B  & 0.5B & 8B & 8B & 8B \\
\# Activated Params & 0.5B & 0.6B & 0.5B & 1B & 1B & 1B \\
\# Training Tokens & 18T & 36T & 4T & 8T & 8T & 8T \\
\midrule
\textbf{General Tasks} \\
\midrule
BBH & 32.27& 41.28& 18.15& 50.34&51.70&60.11 \\
ARC-C & 55.25& 66.10& 46.10& 72.20&67.80&71.53 \\
AGIEval & 30.01& 33.58& 29.29& 38.64&36.52&44.08 \\
HellaSwag & 48.05& 48.88& 44.48& 67.69&67.39&67.43 \\
PIQA & 70.46& 71.33 & 70.29& 77.48&78.84&80.69 \\
MMLU & 49.73& 54.40& 41.76& 58.74&57.83&60.71 \\
CMMLU & 52.10& 51.97& 42.08& 57.68&57.49&64.41 \\
C-Eval & 54.17& 54.57& 44.30& 56.87&58.36&65.98 \\
\midrule
\textbf{Math Tasks} \\
\midrule
GSM8K & 41.32& 60.88& 37.45& 66.41&67.02&72.93 \\
MATH & 18.14& 31.44& 20.66& 37.96&41.98&48.00 \\
CMATH & 52.09& 66.67& 60.75& 74.59&74.13&82.88 \\
\midrule
\textbf{Coding Tasks} \\
\midrule
HumanEval+ & 24.39& 26.83& 29.27& 48.17&50.61&61.59 \\
MBPP+ & 32.80& 38.36& 20.63& 50.26&55.82&62.17 \\
CRUX-O & 14.38& 31.62& 22.56& 35.12&36.31&40.75 \\
\midrule
\textbf{AVG} & 41.08& 48.42& 37.70& 56.58&57.27&63.09 \\
\bottomrule
\end{tabularx}
\label{table:base_main}
\end{table}
For HSA-UltraLong, we developed two variants: a 0.5B dense model and an 8B MoE model with 1B activated parameters. We compared the MoE variant against a standard Transformer-based model (TRM-MoE) with similar parameter count—trained on identical data with matching hyperparameters. The architectures are largely consistent, with only one structural difference: HSA-UltraLong modifies the MoE configuration from 32-expert/2-activated to 64-expert/4-activated with halved expert dimensions. Additionally, HSA-UltraLong uses 16K pretraining context compared to TRM-MoE's 4K. For evaluation, we used the TRM-MoE's 8T-token checkpoint (pre-annealing).

We benchmarked the dense variant against Qwen 2.5-0.5B~\citep{qwen2dot5tech} and Qwen3-0.6B~\citep{qwen3tech}. These comparison models have similar parameter counts but were trained on substantially larger datasets—4.5 times and 9 times our training data volume, respectively.

Our primary evaluation focused on assessing model performance within the pretraining context length across standard benchmarks. Results in Table~\ref{table:base_main} show that the HSA-UltraLong-MoE achieved parity with TRM-MoE in average performance scores, while the dense variant demonstrated only a 3.3-point deficit compared to Qwen 2.5-0.5B, despite having significantly less training data.

Additionally, we evaluated the MoE and Dense models after supervised fine-tuning. The results in Table~\ref{table:inst_main} indicate that while a few general tasks showed no significant performance improvement after supervised fine-tuning, most tasks—particularly math and coding tasks—demonstrated substantial enhancements compared to the base models. Notably, our HSA-UltraLong-MoE achieved scores averaging 1.3 points higher than Qwen3-1.7B (Non-thinking), despite requiring fewer training flops. Similarly, our dense variant performed competitively, scoring only approximately 4 points below Qwen3-0.6B, despite being trained on a dataset merely one-ninth the size.

These findings demonstrate that HSA-UltraLong models maintain their capabilities within standard contexts while extending their effective context length to 16M tokens, further highlighting the superiority of our architectural approach.

\begin{table}[t]
\centering
\renewcommand{\arraystretch}{1.0}
\setlength\tabcolsep{13pt}
\small
\caption{Comparison among HSA-UltraLong-Inst (HSA-UL-Inst) and Qwen3 (Non-thinking) after supervised fine-tuning. All models were evaluated under a unified framework for fair comparison.}
\begin{tabularx}{\linewidth}{lcc|cc@{}}
\toprule
& \makecell{\textbf{Qwen3-Inst}} & \makecell{\textbf{HSA-UL-Inst}} & \makecell{\textbf{Qwen3-Inst}} & 
\makecell{\textbf{HSA-UL-Inst}} \\
\midrule
Architecture & Dense & Dense & Dense & MoE\\
\# Total Params & 0.6B & 0.5B & 1.7B & 8B\\
\# Activated Params & 0.6B & 0.5B & 1.7B & 1B \\
\# Training Tokens & 36T & 4T & 36T & 8T\\
\midrule
\textbf{General Tasks} \\
\midrule
BBH & 42.56 & 26.25 & 59.48 & 57.25\\
MMLU & 45.87 & 42.24 & 63.05 & 61.34\\
CMMLU & 41.64 & 43.33 & 60.84 & 64.06\\
C-Eval & 43.81 & 45.41 & 62.70 & 62.86\\
\midrule
\textbf{Math Tasks} \\
\midrule
GSM8K & 55.65 & 55.42 & 79.00 & 82.94\\
MATH & 45.26 & 40.76 & 64.32 & 61.56\\
MATH500 & 53.00 & 41.00 & 73.20 & 71.00\\
OlympiadBench & 16.89 & 8.74 & 36.30 & 27.85\\
\midrule
\textbf{Coding Tasks} \\
\midrule
HumanEval & 40.24 & 39.63 & 65.24 & 71.95\\
MBPP & 29.20 & 34.40 & 51.00 & 57.00\\
HumanEval+ & 35.37 & 37.20 & 61.59 & 70.73\\
MBPP+ & 34.39 & 39.95 & 59.52 & 65.87\\
CRUX-O & 28.00 & 23.25 & 50.00 & 50.75\\
\midrule
\textbf{Alignment Tasks} \\
\midrule
IFEval Strict Prompt & 55.08 & 33.09 & 64.33 & 63.22\\
\midrule
\textbf{AVG} & 40.50 & 36.48 & 60.76 & 62.03\\
\bottomrule
\end{tabularx}
\label{table:inst_main}
\end{table}

\subsection{Long-context Evaluation}
\begin{figure}[tb]
    \centering
    \begin{subfigure}[b]{0.48\linewidth}
        \centering
        \includegraphics[width=\linewidth]{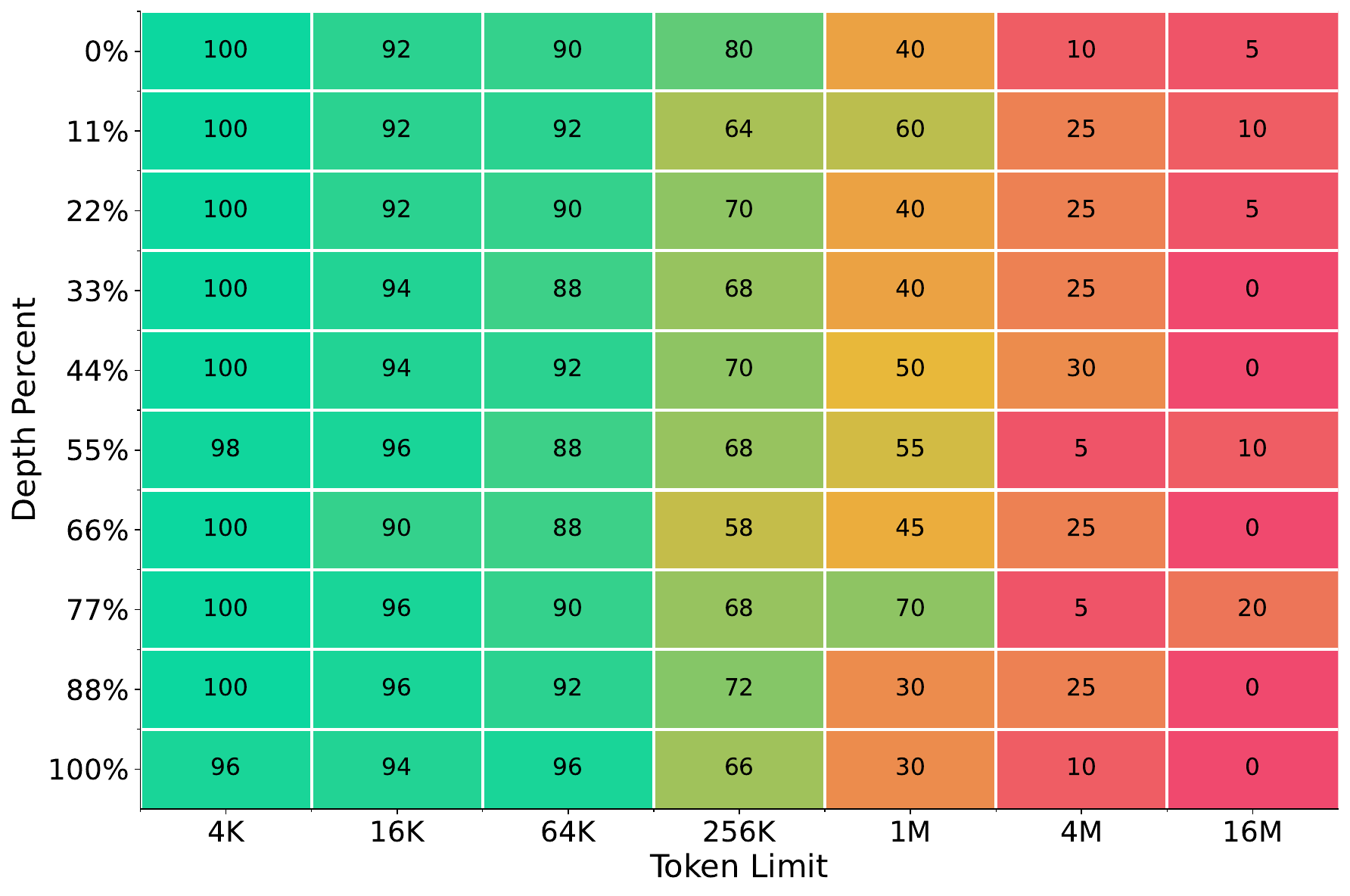}
        \caption{Before long-context mid-training.}
    \end{subfigure}
    \hfill
    \begin{subfigure}[b]{0.48\linewidth}
        \centering
        \includegraphics[width=\linewidth]{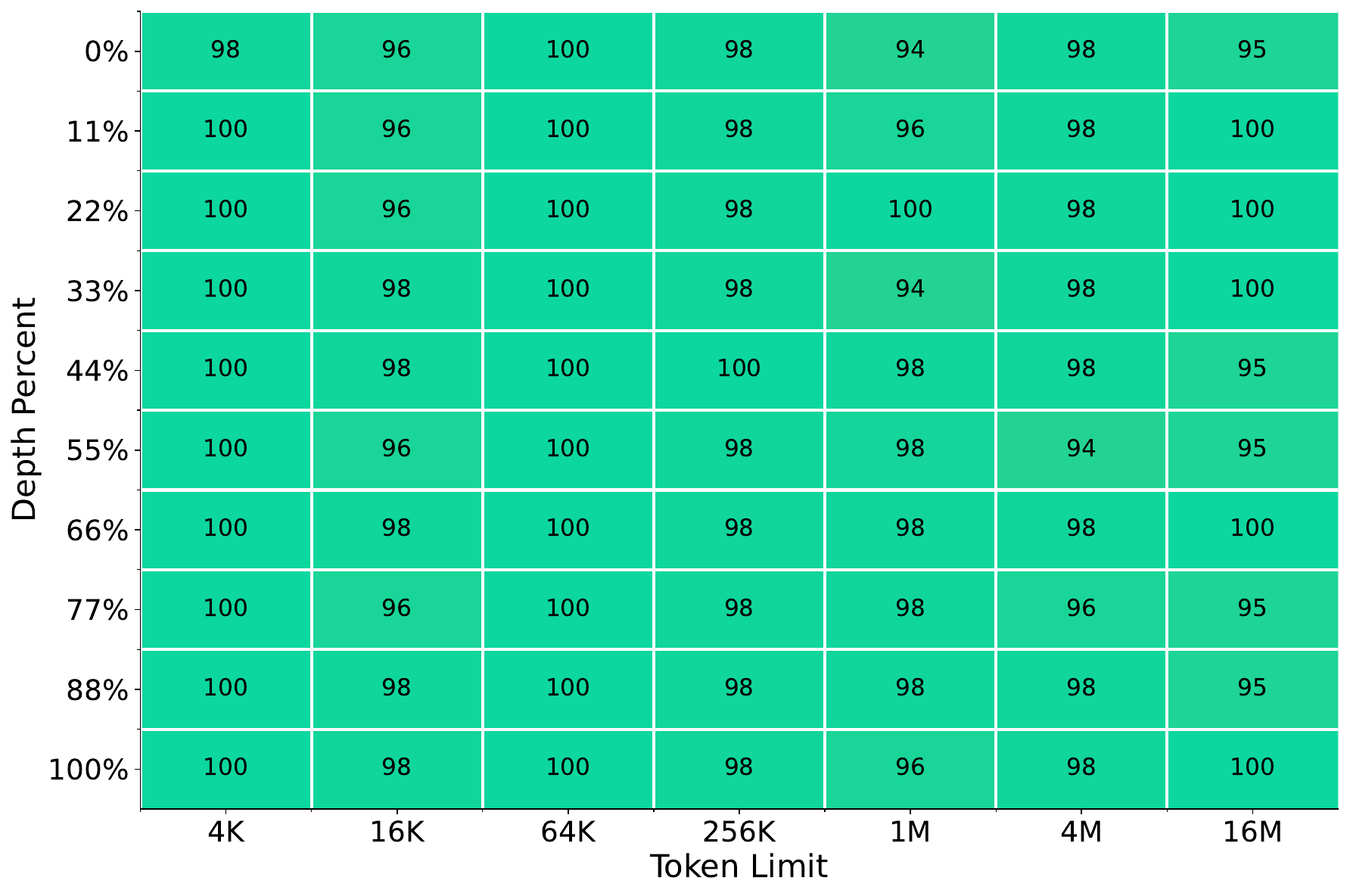}
        \caption{After long-context mid-training.}
    \end{subfigure}
    \\
    \begin{subfigure}[b]{0.48\linewidth}
        \centering
        \includegraphics[width=\linewidth]{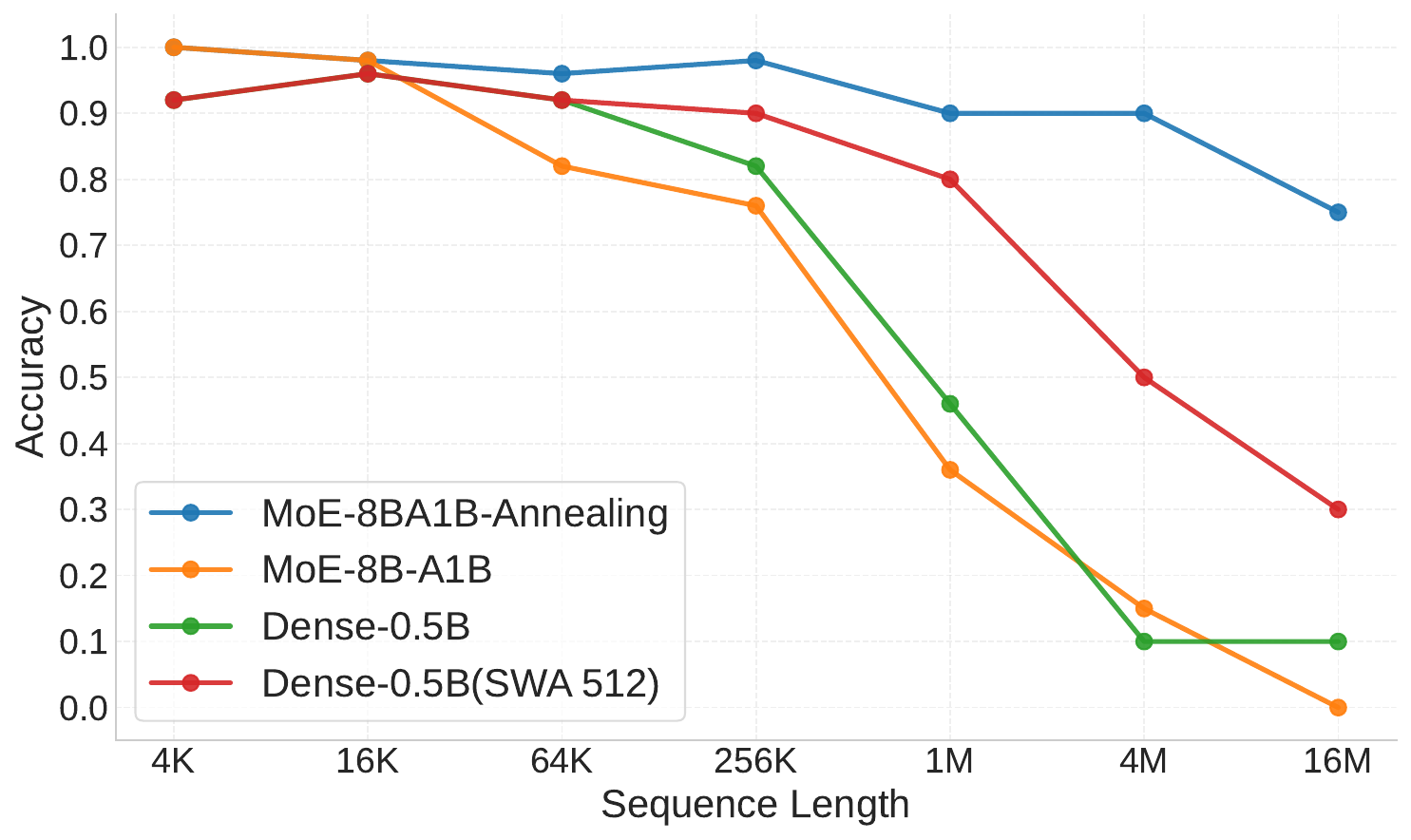}
        \caption{Multi-Query NIAH Task.}
    \end{subfigure}
    \hfill
    \begin{subfigure}[b]{0.48\linewidth}
        \centering
        \includegraphics[width=\linewidth]{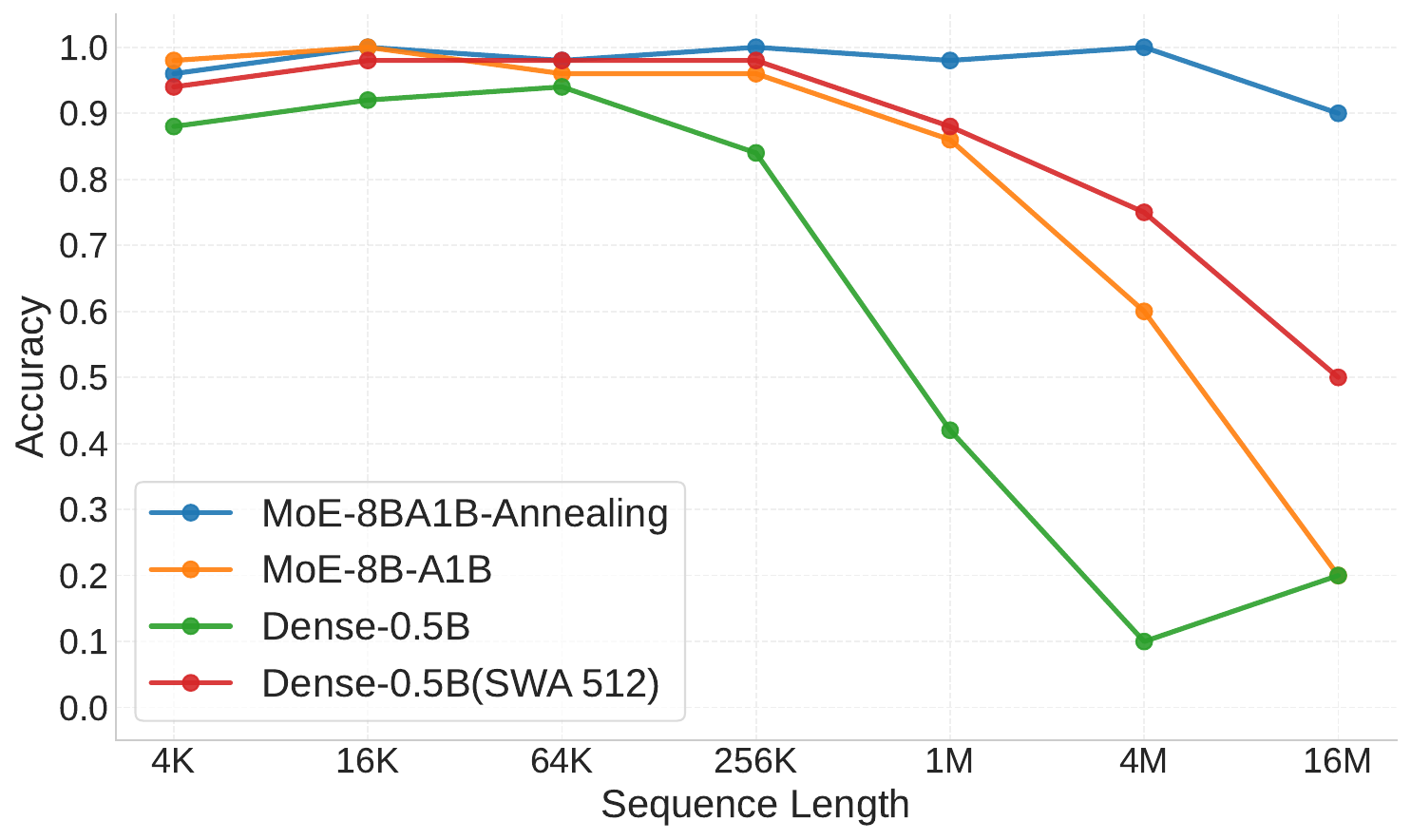}
        \caption{Variable Tracking Task.}
    \end{subfigure}
    \caption{Evaluation of length generalization using the Needle-in-a-Haystack test. (a) and (b) present the results of the HSA-UltraLong-MoE before and after the long-context continued training phase on the Single-NIAH task at various depths. In (c) and (d), we evaluate the performance of different models on the Multi-Query NIAH Task (2 queries, 6 key-value pairs) and the Variable Tracking Task.}
    \label{fig:ruler_eval_line}
\end{figure}

During training, we randomly convert samples into RULER tasks with a 1\% probability by inserting a ``needle'' in a long context and appending the Needle-in-a-Haystack (NIAH) prompt and answer at the end of the sample so the model can follow the NIAH instructions. This modification serves as a probe task to evaluate the model’s extrapolation ability while having minimal impact on training.
The results for RULER tasks are reported in Figure~\ref{fig:ruler_eval_line}, we identify three key findings:

1. \textbf{Effective context length of training data is critical for HSA extrapolation. } As shown in Figure\ref{fig:ruler_eval_line}(a), models pretrained on standard corpora exhibit a progressive decline in retrieval accuracy with longer contexts. This occurs despite a 16K pretraining context window, because the effective context length of the data is often much shorter. In contrast, training on data with longer effective contexts ($>$32K), as in Figure\ref{fig:ruler_eval_line}(b), yields substantially improved extrapolation. This principle underpins the trends in Figures~\ref{fig:ruler_eval_line}(c) and (d).

2. \textbf{A seesaw effect exists between HSA and Sliding Window Attention. } Figures~\ref{fig:ruler_eval_line}(c) and (d) indicate that a smaller SWA window (512) during continued pretraining leads to better HSA extrapolation than a larger window (4K). Given that training from scratch with a 4K window fails to develop extrapolative HSA, we conclude that larger SWA windows impair HSA's long-range generalization. We posit that HSA learns a form of retrieval-based extrapolation. Large SWA windows handle most short-range dependencies inherently, reducing the incentive for HSA to learn them and thus weakening its ability to generalize to longer sequences.

3. \textbf{HSA capability scales with parameter size in reasoning-retrieval tasks.} While MoE-8B-A1B and Dense-0.5B exhibit comparable performance on the pure retrieval task (MQ-NIAH; Figure~\ref{fig:ruler_eval_line}(c)), MoE-8B-A1B consistently outperforms Dense-0.5B in the variable-tracking task (Figure~\ref{fig:ruler_eval_line}(d)), demonstrating that larger models better support joint reasoning and retrieval.

\subsection{Training/Inference Efficiency}
\begin{figure}[tb]
    \centering
    \begin{subfigure}[b]{0.48\linewidth}
        \centering
        \includegraphics[width=\linewidth]{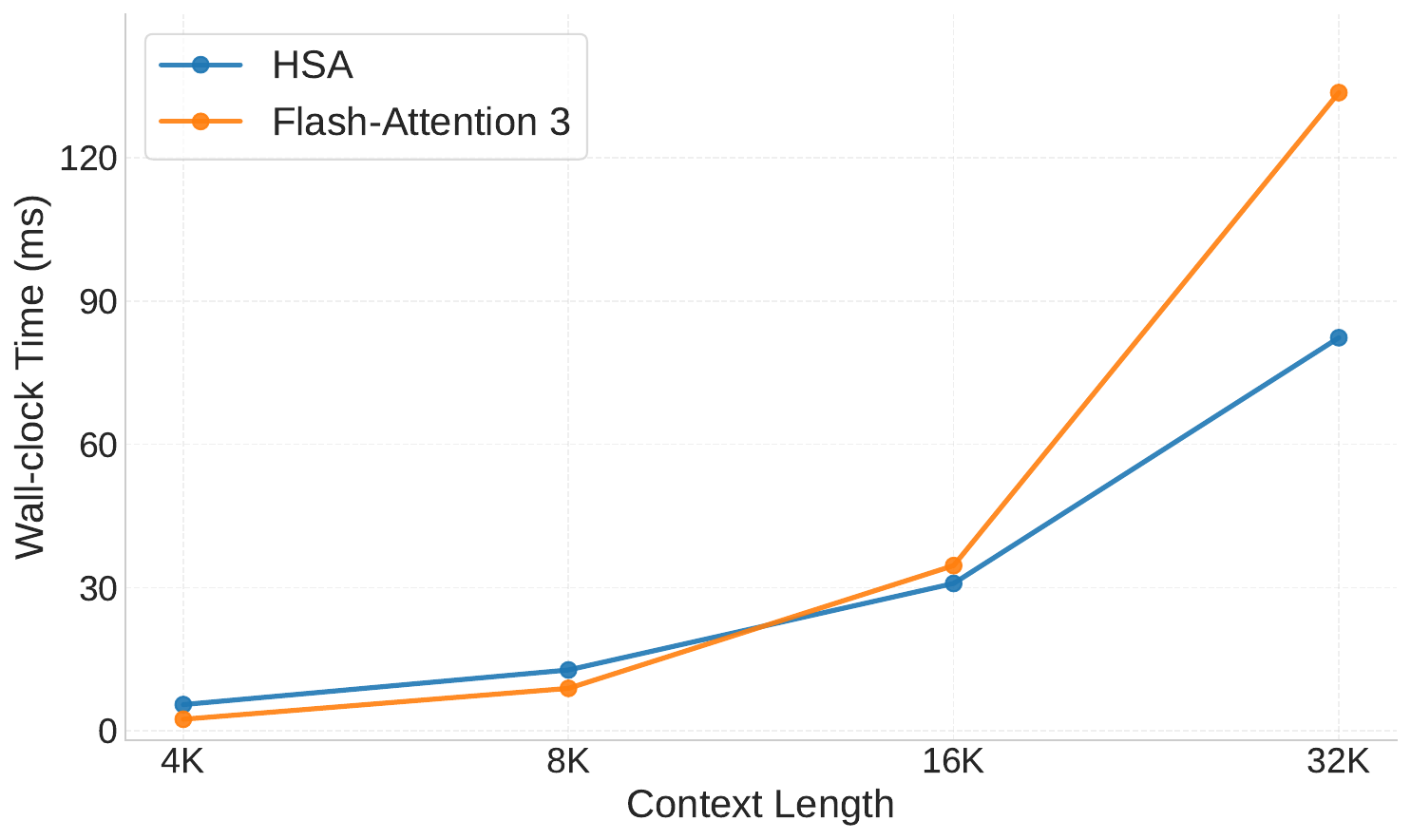}
        \caption{Training efficiency.}
        \label{fig:training_efficiency}
    \end{subfigure}
    % \hfill
    \begin{subfigure}[b]{0.48\linewidth}
        \centering
        \includegraphics[width=\linewidth]{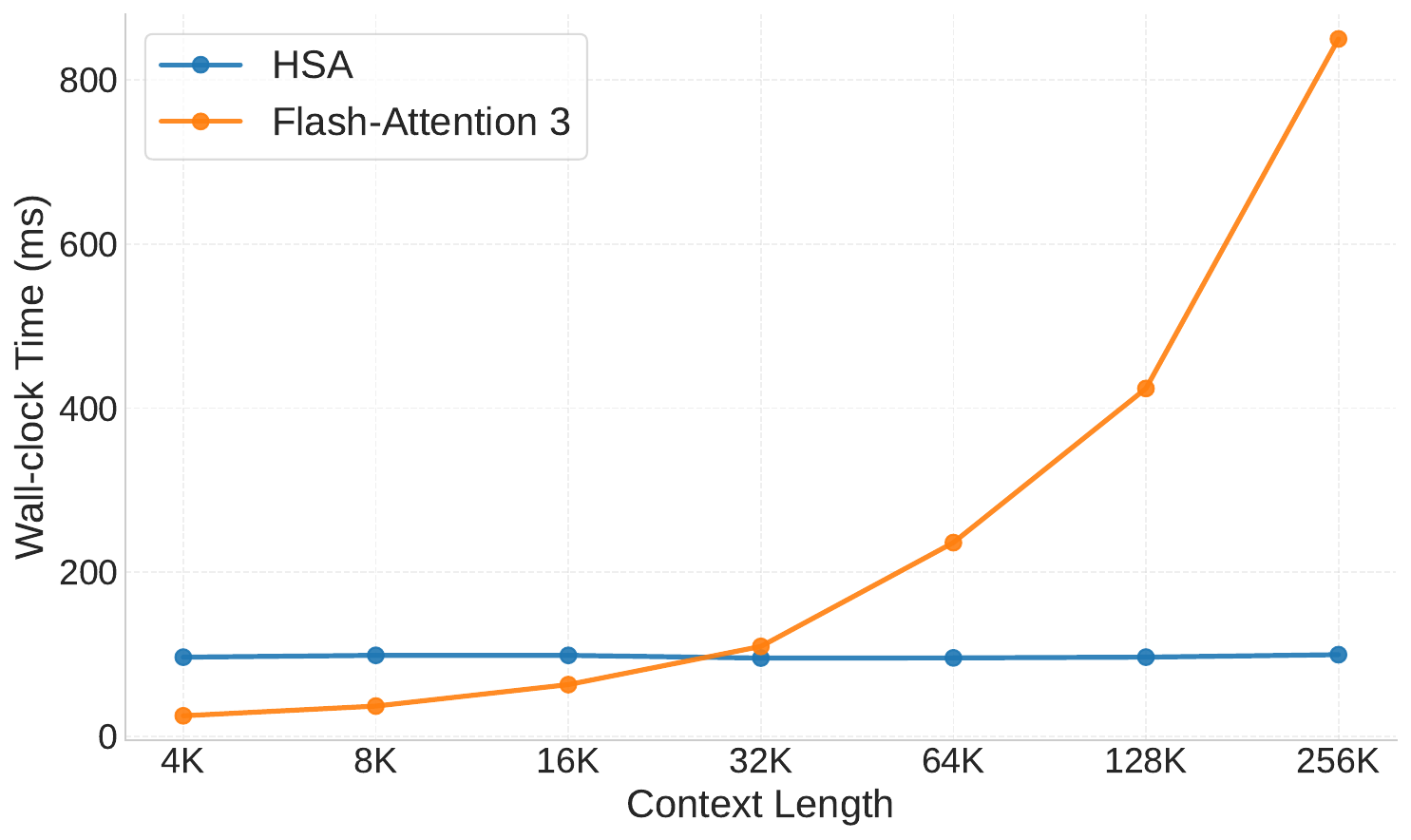}
        \caption{Inference efficiency.}
        \label{fig:inference_efficiency}
    \end{subfigure}
    \caption{Comparison of training/inference efficiency between HSA kernel and Flash-attention 3.}
    \label{fig:efficiency}
\end{figure}

To further evaluate the efficiency of the sparse attention module, we benchmark the HSA operator against the FlashAttention-3~\citep{shah2024flashattention} operator on H800 for both training and inference, with HSA implemented using TileLang~\citep{wang2025tilelangcomposabletiledprogramming}. As shown in Figure~\ref{fig:efficiency}, at shorter sequence lengths, FlashAttention-3 still leads in both training and inference, and HSA only gains an advantage with longer contexts. We attribute this to two factors: (1) the sparsity in HSA causes the kernel to incur more memory accesses compared to FlashAttention-3; and (2) FlashAttention-3 is implemented in CUDA, enabling it to better leverage the features of the Hopper architecture.
\section{Conclusion}
Although HSA has shown promising extrapolation capabilities, several challenges remain:
\begin{itemize}
    \item The HSA/SWA seesaw problem. After training on short SFT data, extrapolation can degrade. The main reason is that an excessively long sliding-window attention reduces the need for HSA to learn short-range dependencies, which in turn hampers its ability to extrapolate to long-range dependencies.
    \item The head ratio constraint. HSA currently requires a 16:1 ratio of query heads to key–value heads, creating a severe information bottleneck. Future work should pursue kernel-level optimizations to alleviate this constraint.
    \item 
    When sequences are short, training and inference show no clear advantage over FlashAttention-3; further kernel-level optimizations are needed to improve efficiency.
\end{itemize}
Despite these limitations, HSA-UltraLong presents a highly promising paradigm for long-context processing. The core insight of HSA is to \textit{perform attention chunk by chunk and fuse the results via retrieval scores}, rather than selecting chunks and then concatenating them for attention. The experimental results provide a meaningful step toward effectively handling infinite-long context, advancing progress on long-term memory in machines.

\newpage
%%%%%%%%%%%%%%%%%%%%%%%%%%%%%%%%%%%%%%%%%%%%%%%%%%%%%%%%%%%%%%%%%%
%%%%%%%%%%%%%%%%%%%%%%%%%%%%%%%%%%%%%%%%%%%%%%%%%%%%%%%%%%%%%%%%%%
%%%%%%
%%%%%%%%%%%%%%%%%%%%%%%%%%%%%%%%%%%%%%%%%%%%%%%%%%%%%%%%%%%

%\clearpage
%\appendix

%%reference
%\clearpage
\bibliographystyle{antgroup}
\bibliography{ref/reference}

@inproceedings{
mohtashami2023randomaccess,
title={Random-Access Infinite Context Length for Transformers},
author={Amirkeivan Mohtashami and Martin Jaggi},
booktitle={Thirty-seventh Conference on Neural Information Processing Systems},
year={2023},
url={https://openreview.net/forum?id=7eHn64wOVy}
}

@inproceedings{
hu2025efficient,
title={Efficient Length-Generalizable Attention via Causal Retrieval for Long-Context Language Modeling},
author={Xiang Hu and Zhihao Teng and Jun Zhao and Wei Wu and Kewei Tu},
booktitle={Forty-second International Conference on Machine Learning},
year={2025},
url={https://openreview.net/forum?id=6HVcoIbZoC}
}

@inproceedings{
hu2025hardwarealigned,
title={Hardware-aligned Hierarchical Sparse Attention for Efficient Long-term Memory Access},
author={Xiang Hu and Jiaqi Leng and Jun Zhao and Kewei Tu and Wei Wu},
booktitle={The Thirty-ninth Annual Conference on Neural Information Processing Systems},
year={2025},
url={https://openreview.net/forum?id=dIHSZTx9Lu}
}

@misc{leng2025understandingimprovinglengthgeneralization,
      title={Understanding and Improving Length Generalization in Hierarchical Sparse Attention Models}, 
      author={Jiaqi Leng and Xiang Hu and Junxiong Wang and Jianguo Li and Wei Wu and Yucheng Lu},
      year={2025},
      eprint={2510.17196},
      archivePrefix={arXiv},
      primaryClass={cs.CL},
      url={https://arxiv.org/abs/2510.17196}, 
}

@inproceedings{DBLP:conf/acl/WuT24,
  author       = {Haoyi Wu and
                  Kewei Tu},
  editor       = {Lun{-}Wei Ku and
                  Andre Martins and
                  Vivek Srikumar},
  title        = {Layer-Condensed {KV} Cache for Efficient Inference of Large Language
                  Models},
  booktitle    = {Proceedings of the 62nd Annual Meeting of the Association for Computational
                  Linguistics (Volume 1: Long Papers), {ACL} 2024, Bangkok, Thailand,
                  August 11-16, 2024},
  pages        = {11175--11188},
  publisher    = {Association for Computational Linguistics},
  year         = {2024},
  url          = {https://doi.org/10.18653/v1/2024.acl-long.602},
  doi          = {10.18653/V1/2024.ACL-LONG.602},
  bibsource    = {dblp computer science bibliography, https://dblp.org}
}

@article{rubin-berant-2024-retrieval,
  title={Retrieval-Pretrained Transformer: Long-range Language Modeling with Self-retrieval},
  author={Rubin, Ohad and Berant, Jonathan},
  journal={Transactions of the Association for Computational Linguistics},
  volume={12},
  pages={1197--1213},
  year={2024},
  publisher={MIT Press 255 Main Street, 9th Floor, Cambridge, Massachusetts 02142, USA~…}
}

@misc{deepseekai2024deepseekv3,
    title={DeepSeek-V3 Technical Report},
    author={DeepSeek-AI and Aixin Liu and Bei Feng and Bing Xue and Bingxuan Wang and others},
    year={2024},
    eprint={2412.19437},
    archivePrefix={arXiv},
    primaryClass={cs.CL}
}

@article{DBLP:journals/corr/abs-2408-15664,
  author       = {Lean Wang and
                  Huazuo Gao and
                  Chenggang Zhao and
                  Xu Sun and
                  Damai Dai},
  title        = {Auxiliary-Loss-Free Load Balancing Strategy for Mixture-of-Experts},
  journal      = {CoRR},
  volume       = {abs/2408.15664},
  year         = {2024},
  url          = {https://doi.org/10.48550/arXiv.2408.15664},
  doi          = {10.48550/ARXIV.2408.15664},
  eprinttype    = {arXiv},
  eprint       = {2408.15664},
  bibsource    = {dblp computer science bibliography, https://dblp.org}
}

@misc{lingteam2025activationboostedscalinggeneral,
      title={Every Activation Boosted: Scaling General Reasoner to 1 Trillion Open Language Foundation}, 
      author={Ling-Team and Ang Li and Ben Liu and Binbin Hu and Bing Li and Bingwei Zeng and Borui Ye and others},
      year={2025},
      eprint={2510.22115},
      archivePrefix={arXiv},
      primaryClass={cs.CL},
      url={https://arxiv.org/abs/2510.22115}, 
}

@inproceedings{NIPS2017_3f5ee243,
 author = {Vaswani, Ashish and Shazeer, Noam and Parmar, Niki and others},
 pages = {},
 publisher = {Curran Associates, Inc.},
 title = {Attention is All you Need},
 url = {https://proceedings.neurips.cc/paper_files/paper/2017/file/3f5ee243547dee91fbd053c1c4a845aa-Paper.pdf},
 volume = {30},
 year = {2017}
}

@inproceedings{
hsieh2024ruler,
title={{RULER}: What{\textquoteright}s the Real Context Size of Your Long-Context Language Models?},
author={Cheng-Ping Hsieh and Simeng Sun and Samuel Kriman and Shantanu Acharya and Dima Rekesh and Fei Jia and Boris Ginsburg},
booktitle={First Conference on Language Modeling},
year={2024},
url={https://openreview.net/forum?id=kIoBbc76Sy}
}

@misc{kuratov2024babilongtestinglimitsllms,
      title={BABILong: Testing the Limits of LLMs with Long Context Reasoning-in-a-Haystack}, 
      author={Yuri Kuratov and Aydar Bulatov and Petr Anokhin and Ivan Rodkin and Dmitry Sorokin and Artyom Sorokin and Mikhail Burtsev},
      year={2024},
      eprint={2406.10149},
      archivePrefix={arXiv},
      primaryClass={cs.CL},
      url={https://arxiv.org/abs/2406.10149}, 
}

@misc{dehghani2023scalingvisiontransformers22,
      title={Scaling Vision Transformers to 22 Billion Parameters}, 
      author={Mostafa Dehghani and Josip Djolonga and Basil Mustafa and others},
      year={2023},
      eprint={2302.05442},
      archivePrefix={arXiv},
      primaryClass={cs.CV},
      url={https://arxiv.org/abs/2302.05442}, 
}

@misc{wortsman2023smallscale,
    title={Small-scale proxies for large-scale Transformer training instabilities},
    author={Mitchell Wortsman and Peter J. Liu and Lechao Xiao and Katie Everett and others},
    year={2023},
    eprint={2309.14322},
    archivePrefix={arXiv},
    primaryClass={cs.LG}
}

@article{hendrycks2020mmlu,
  title={{Measuring Massive Multitask Language Understanding}},
  author={Hendrycks, Dan and Burns, Collin and Basart, Steven and Zou, Andy and Mazeika, Mantas and Song, Dawn and Steinhardt, Jacob},
  journal={arXiv preprint arXiv:2009.03300},
  year={2020}
}

@article{suzgun2022bbh,
  title={{Challenging Big-Bench Tasks and Whether Chain-of-Thought Can Solve Them}},
  author={Suzgun, Mirac and Scales, Nathan and Sch{\"a}rli, Nathanael and Gehrmann, Sebastian and Tay, Yi and Chung, Hyung Won and Chowdhery, Aakanksha and Le, Quoc V and Chi, Ed H and Zhou, Denny and others},
  journal={arXiv preprint arXiv:2210.09261},
  year={2022}
}

@article{huang2023ceval,
  title={{C-Eval: A Multi-Level Multi-Discipline Chinese Evaluation Suite for Foundation Models}},
  author={Huang, Yuzhen and Bai, Yuzhuo and Zhu, Zhihao and Zhang, Junlei and Zhang, Jinghan and Su, Tangjun and Liu, Junteng and Lv, Chuancheng and others},
  journal={Advances in Neural Information Processing Systems},
  volume={36},
  pages={62991--63010},
  year={2023}
}

@article{li2023cmmlu,
  title={{CMMLU: Measuring Massive Multitask Language Understanding in Chinese}},
  author={Li, Haonan and Zhang, Yixuan and Koto, Fajri and Yang, Yifei and Zhao, Hai and Gong, Yeyun and Duan, Nan and Baldwin, Timothy},
  journal={arXiv preprint arXiv:2306.09212},
  year={2023}
}

@article{cobbe2021gsm8k,
  title={{Training Verifiers to Solve Math Word Problems}},
  author={Cobbe, Karl and Kosaraju, Vineet and Bavarian, Mohammad and Chen, Mark and Jun, Heewoo and Kaiser, Lukasz and Plappert, Matthias and Tworek, Jerry and Hilton, Jacob and Nakano, Reiichiro and others},
  journal={arXiv preprint arXiv:2110.14168},
  year={2021}
}

@article{hendrycks2021math,
  title={{Measuring Mathematical Problem Solving with the Math Dataset}},
  author={Hendrycks, Dan and Burns, Collin and Kadavath, Saurav and Arora, Akul and Basart, Steven and Tang, Eric and Song, Dawn and Steinhardt, Jacob},
  journal={arXiv preprint arXiv:2103.03874},
  year={2021}
}

@inproceedings{lightman2023math500,
  title={{Let's Verify Step by Step}},
  author={Lightman, Hunter and Kosaraju, Vineet and Burda, Yuri and Edwards, Harrison and Baker, Bowen and Lee, Teddy and Leike, Jan and Schulman, John and Sutskever, Ilya and Cobbe, Karl},
  booktitle={The Twelfth International Conference on Learning Representations},
  year={2023}
}

@article{he2024olympiadbench,
  title={{OlympiadBench: A Challenging Benchmark for Promoting AGI with Olympiad-Level Bilingual Multimodal Scientific Problems}},
  author={He, Chaoqun and Luo, Renjie and Bai, Yuzhuo and Hu, Shengding and Thai, Zhen Leng and Shen, Junhao and Hu, Jinyi and Han, Xu and Huang, Yujie and Zhang, Yuxiang and others},
  journal={arXiv preprint arXiv:2402.14008},
  year={2024}
}

@article{liu2023evalplus,
  title={{Is Your Code Generated by ChatGPT Really Correct? Rigorous Evaluation of Large Language Models for Code Generation}},
  author={Liu, Jiawei and Xia, Chunqiu Steven and Wang, Yuyao and Zhang, Lingming},
  journal={Advances in Neural Information Processing Systems},
  volume={36},
  pages={21558--21572},
  year={2023}
}

@article{gu2024cruxeval,
  title={{CruxEval: A Benchmark for Code Reasoning, Understanding and Execution}},
  author={Gu, Alex and Rozi{\`e}re, Baptiste and Leather, Hugh and Solar-Lezama, Armando and Synnaeve, Gabriel and Wang, Sida I},
  journal={arXiv preprint arXiv:2401.03065},
  year={2024}
}

@article{zhou2023ifeval,
  title={{Instruction-Following Evaluation for Large Language Models}},
  author={Zhou, Jeffrey and Lu, Tianjian and Mishra, Swaroop and Brahma, Siddhartha and Basu, Sujoy and Luan, Yi and Zhou, Denny and Hou, Le},
  journal={arXiv preprint arXiv:2311.07911},
  year={2023}
}

@inproceedings{piqa,
  author       = {Yonatan Bisk and
                  Rowan Zellers and
                  Ronan Le Bras and
                  Jianfeng Gao and
                  Yejin Choi},
  title        = {{PIQA:} Reasoning about Physical Commonsense in Natural Language},
  booktitle    = {The Thirty-Fourth {AAAI} Conference on Artificial Intelligence, {AAAI}
                  2020, The Thirty-Second Innovative Applications of Artificial Intelligence
                  Conference, {IAAI} 2020, The Tenth {AAAI} Symposium on Educational
                  Advances in Artificial Intelligence, {EAAI} 2020, New York, NY, USA,
                  February 7-12, 2020},
  pages        = {7432--7439},
  publisher    = {{AAAI} Press},
  year         = {2020},
  url          = {https://doi.org/10.1609/aaai.v34i05.6239},
  doi          = {10.1609/AAAI.V34I05.6239},
  bibsource    = {dblp computer science bibliography, https://dblp.org}
}

@inproceedings{agieval,
  author       = {Wanjun Zhong and
                  Ruixiang Cui and
                  Yiduo Guo and
                  Yaobo Liang and
                  Shuai Lu and
                  Yanlin Wang and
                  Amin Saied and
                  Weizhu Chen and
                  Nan Duan},
  editor       = {Kevin Duh and
                  Helena G{\'{o}}mez{-}Adorno and
                  Steven Bethard},
  title        = {AGIEval: {A} Human-Centric Benchmark for Evaluating Foundation Models},
  booktitle    = {Findings of the Association for Computational Linguistics: {NAACL}
                  2024, Mexico City, Mexico, June 16-21, 2024},
  pages        = {2299--2314},
  publisher    = {Association for Computational Linguistics},
  year         = {2024},
  url          = {https://doi.org/10.18653/v1/2024.findings-naacl.149},
  doi          = {10.18653/V1/2024.FINDINGS-NAACL.149},
  bibsource    = {dblp computer science bibliography, https://dblp.org}
}

@inproceedings{hellaswag,
  author       = {Rowan Zellers and
                  Ari Holtzman and
                  Yonatan Bisk and
                  Ali Farhadi and
                  Yejin Choi},
  editor       = {Anna Korhonen and
                  David R. Traum and
                  Llu{\'{\i}}s M{\`{a}}rquez},
  title        = {HellaSwag: Can a Machine Really Finish Your Sentence?},
  booktitle    = {Proceedings of the 57th Conference of the Association for Computational
                  Linguistics, {ACL} 2019, Florence, Italy, July 28- August 2, 2019,
                  Volume 1: Long Papers},
  pages        = {4791--4800},
  publisher    = {Association for Computational Linguistics},
  year         = {2019},
  url          = {https://doi.org/10.18653/v1/p19-1472},
  doi          = {10.18653/V1/P19-1472},
  bibsource    = {dblp computer science bibliography, https://dblp.org}
}

@article{arc,
  author       = {Sumithra Bhakthavatsalam and
                  Daniel Khashabi and
                  Tushar Khot and
                  Bhavana Dalvi Mishra and
                  Kyle Richardson and
                  Ashish Sabharwal and
                  Carissa Schoenick and
                  Oyvind Tafjord and
                  Peter Clark},
  title        = {Think you have Solved Direct-Answer Question Answering? Try ARC-DA,
                  the Direct-Answer {AI2} Reasoning Challenge},
  journal      = {CoRR},
  volume       = {abs/2102.03315},
  year         = {2021},
  url          = {https://arxiv.org/abs/2102.03315},
  eprinttype    = {arXiv},
  eprint       = {2102.03315},
  bibsource    = {dblp computer science bibliography, https://dblp.org}
}

@inproceedings{yuan-etal-2025-native,
  title={Native sparse attention: Hardware-aligned and natively trainable sparse attention},
  author={Yuan, Jingyang and Gao, Huazuo and Dai, Damai and Luo, Junyu and Zhao, Liang and Zhang, Zhengyan and Xie, Zhenda and Wei, Yuxing and Wang, Lean and Xiao, Zhiping and others},
  booktitle={Proceedings of the 63rd Annual Meeting of the Association for Computational Linguistics (Volume 1: Long Papers)},
  pages={23078--23097},
  year={2025}
}

@inproceedings{
lu2025moba,
title={Mo{BA}: Mixture of Block Attention for Long-Context {LLM}s},
author={Enzhe Lu and Zhejun Jiang and Jingyuan Liu and Yulun Du and Tao Jiang and Chao Hong and Shaowei Liu and Weiran He and Enming Yuan and Yuzhi Wang and Zhiqi Huang and Huan Yuan and Suting Xu and Xinran Xu and Guokun Lai and Yanru Chen and Huabin Zheng and Junjie Yan and Jianlin Su and Yuxin Wu and Yutao Zhang and Zhilin Yang and Xinyu Zhou and Mingxing Zhang and Jiezhong Qiu},
booktitle={The Thirty-ninth Annual Conference on Neural Information Processing Systems},
year={2025},
url={https://openreview.net/forum?id=RlqYCpTu1P}
}

@article{DBLP:journals/ijon/SuALPBL24,
  author       = {Jianlin Su and
                  Murtadha H. M. Ahmed and
                  Yu Lu and
                  Shengfeng Pan and
                  Wen Bo and
                  Yunfeng Liu},
  title        = {RoFormer: Enhanced transformer with Rotary Position Embedding},
  journal      = {Neurocomputing},
  volume       = {568},
  pages        = {127063},
  year         = {2024},
  url          = {https://doi.org/10.1016/j.neucom.2023.127063},
  doi          = {10.1016/J.NEUCOM.2023.127063},
  bibsource    = {dblp computer science bibliography, https://dblp.org}
}

@article{humaneval,
  author       = {Mark Chen and
                  Jerry Tworek and
                  Heewoo Jun and
                  Qiming Yuan and
                  Henrique Pond{\'{e}} de Oliveira Pinto and
                  others},
  title        = {Evaluating Large Language Models Trained on Code},
  journal      = {CoRR},
  volume       = {abs/2107.03374},
  year         = {2021},
  url          = {https://arxiv.org/abs/2107.03374},
  eprinttype    = {arXiv},
  eprint       = {2107.03374},
  bibsource    = {dblp computer science bibliography, https://dblp.org}
}

@article{mbpp,
  author       = {Ning Tao and
                  Anthony Ventresque and
                  Vivek Nallur and
                  Takfarinas Saber},
  title        = {Enhancing Program Synthesis with Large Language Models Using Many-Objective
                  Grammar-Guided Genetic Programming},
  journal      = {Algorithms},
  volume       = {17},
  number       = {7},
  pages        = {287},
  year         = {2024},
  url          = {https://doi.org/10.3390/a17070287},
  doi          = {10.3390/A17070287},
  bibsource    = {dblp computer science bibliography, https://dblp.org}
}

@article{cmath,
  author       = {Tianwen Wei and
                  Jian Luan and
                  Wei Liu and
                  Shuang Dong and
                  Bin Wang},
  title        = {{CMATH:} Can Your Language Model Pass Chinese Elementary School Math
                  Test?},
  journal      = {CoRR},
  volume       = {abs/2306.16636},
  year         = {2023},
  url          = {https://doi.org/10.48550/arXiv.2306.16636},
  doi          = {10.48550/ARXIV.2306.16636},
  eprinttype    = {arXiv},
  eprint       = {2306.16636},
  bibsource    = {dblp computer science bibliography, https://dblp.org}
}

@article{wu2025grove,
  title={Grove moe: Towards efficient and superior moe llms with adjugate experts},
  author={Wu, Haoyuan and Chen, Haoxing and Chen, Xiaodong and Zhou, Zhanchao and Chen, Tieyuan and Zhuang, Yihong and Lu, Guoshan and Huang, Zenan and Zhao, Junbo and Liu, Lin and others},
  journal={arXiv preprint arXiv:2508.07785},
  year={2025}
}

@article{qwen2dot5tech,
  title={Qwen2.5 technical report},
  author={Yang, An and Yang, Baosong and Zhang, Beichen and Hui, Binyuan and Zheng, Bo and Yu, Bowen and Li, Chengyuan and Liu, Dayiheng and Huang, Fei and Wei, Haoran and others},
  journal={arXiv preprint arXiv:2412.15115},
  year={2024}
}

@article{qwen3tech,
  title={{Qwen3 Technical Report}},
  author={Yang, An and Li, Anfeng and Yang, Baosong and Zhang, Beichen and Hui, Binyuan and Zheng, Bo and Yu, Bowen and Gao, Chang and Huang, Chengen and Lv, Chenxu and others},
  journal={arXiv preprint arXiv:2505.09388},
  year={2025}
}

@inproceedings{
shah2024flashattention,
title={FlashAttention-3: Fast and Accurate Attention with Asynchrony and Low-precision},
author={Jay Shah and Ganesh Bikshandi and Ying Zhang and Vijay Thakkar and Pradeep Ramani and Tri Dao},
booktitle={The Thirty-eighth Annual Conference on Neural Information Processing Systems},
year={2024},
url={https://openreview.net/forum?id=tVConYid20}
}

@misc{wang2025tilelangcomposabletiledprogramming,
      title={TileLang: A Composable Tiled Programming Model for AI Systems}, 
      author={Lei Wang and Yu Cheng and Yining Shi and Zhengju Tang and Zhiwen Mo and Wenhao Xie and Lingxiao Ma and Yuqing Xia and Jilong Xue and Fan Yang and Zhi Yang},
      year={2025},
      eprint={2504.17577},
      archivePrefix={arXiv},
      primaryClass={cs.LG},
      url={https://arxiv.org/abs/2504.17577}, 
}

@article{Cowan2008WhatAT,
  title={What are the differences between long-term, short-term, and working memory?},
  author={Nelson Cowan},
  journal={Progress in brain research},
  year={2008},
  volume={169},
  pages={
          323-38
        },
  url={https://api.semanticscholar.org/CorpusID:205921304}
}

@article{DBLP:journals/corr/abs-2312-00752,
  author       = {Albert Gu and
                  Tri Dao},
  title        = {Mamba: Linear-Time Sequence Modeling with Selective State Spaces},
  journal      = {CoRR},
  volume       = {abs/2312.00752},
  year         = {2023},
  url          = {https://doi.org/10.48550/arXiv.2312.00752},
  doi          = {10.48550/ARXIV.2312.00752},
  eprinttype    = {arXiv},
  eprint       = {2312.00752},
  bibsource    = {dblp computer science bibliography, https://dblp.org}
}

@inproceedings{DBLP:conf/icml/KatharopoulosV020,
  author       = {Angelos Katharopoulos and
                  Apoorv Vyas and
                  Nikolaos Pappas and
                  Fran{\c{c}}ois Fleuret},
  title        = {Transformers are RNNs: Fast Autoregressive Transformers with Linear
                  Attention},
  booktitle    = {Proceedings of the 37th International Conference on Machine Learning,
                  {ICML} 2020, 13-18 July 2020, Virtual Event},
  series       = {Proceedings of Machine Learning Research},
  volume       = {119},
  pages        = {5156--5165},
  publisher    = {{PMLR}},
  year         = {2020},
  url          = {http://proceedings.mlr.press/v119/katharopoulos20a.html},
  bibsource    = {dblp computer science bibliography, https://dblp.org}
}

@inproceedings{DBLP:conf/icml/DaoG24,
  author       = {Tri Dao and
                  Albert Gu},
  title        = {Transformers are SSMs: Generalized Models and Efficient Algorithms
                  Through Structured State Space Duality},
  booktitle    = {Forty-first International Conference on Machine Learning, {ICML} 2024,
                  Vienna, Austria, July 21-27, 2024},
  publisher    = {OpenReview.net},
  year         = {2024},
  url          = {https://openreview.net/forum?id=ztn8FCR1td},
  bibsource    = {dblp computer science bibliography, https://dblp.org}
}

@inproceedings{DBLP:conf/iclr/YangKH25,
  author       = {Songlin Yang and
                  Jan Kautz and
                  Ali Hatamizadeh},
  title        = {Gated Delta Networks: Improving Mamba2 with Delta Rule},
  booktitle    = {The Thirteenth International Conference on Learning Representations,
                  {ICLR} 2025, Singapore, April 24-28, 2025},
  publisher    = {OpenReview.net},
  year         = {2025},
  url          = {https://openreview.net/forum?id=r8H7xhYPwz},
  bibsource    = {dblp computer science bibliography, https://dblp.org}
}

@article{Beltagy2020Longformer,
  title={Longformer: The Long-Document Transformer},
  author={Iz Beltagy and Matthew E. Peters and Arman Cohan},
  journal={arXiv:2004.05150},
  year={2020},
}

@inproceedings{
shazeer2017,
title={ Outrageously Large Neural Networks: The Sparsely-Gated Mixture-of-Experts Layer},
author={Noam Shazeer and *Azalia Mirhoseini and *Krzysztof Maziarz and Andy Davis and Quoc Le and Geoffrey Hinton and Jeff Dean},
booktitle={International Conference on Learning Representations},
year={2017},
url={https://openreview.net/forum?id=B1ckMDqlg}
}

@article{achiam2023gpt,
  title={Gpt-4 technical report},
  author={Achiam, Josh and Adler, Steven and Agarwal, Sandhini and Ahmad, Lama and Akkaya, Ilge and Aleman, Florencia Leoni and Almeida, Diogo and Altenschmidt, Janko and Altman, Sam and Anadkat, Shyamal and others},
  journal={arXiv preprint arXiv:2303.08774},
  year={2023}
}

@article{touvron2023llama,
  title={Llama 2: Open foundation and fine-tuned chat models},
  author={Touvron, Hugo and Martin, Louis and Stone, Kevin and Albert, Peter and Almahairi, Amjad and Babaei, Yasmine and Bashlykov, Nikolay and Batra, Soumya and Bhargava, Prajjwal and Bhosale, Shruti and others},
  journal={arXiv preprint arXiv:2307.09288},
  year={2023}
}

@inproceedings{DBLP:conf/nips/BrownMRSKDNSSAA20,
  author       = {Tom B. Brown and
                  Benjamin Mann and
                  Nick Ryder and
                  Melanie Subbiah and
                  others},
  editor       = {Hugo Larochelle and
                  Marc'Aurelio Ranzato and
                  Raia Hadsell and
                  Maria{-}Florina Balcan and
                  Hsuan{-}Tien Lin},
  title        = {Language Models are Few-Shot Learners},
  booktitle    = {Advances in Neural Information Processing Systems 33: Annual Conference
                  on Neural Information Processing Systems 2020, NeurIPS 2020, December
                  6-12, 2020, virtual},
  year         = {2020},
  url          = {https://proceedings.neurips.cc/paper/2020/hash/1457c0d6bfcb4967418bfb8ac142f64a-Abstract.html},
  bibsource    = {dblp computer science bibliography, https://dblp.org}
}

\end{document}